%% file: iclr2023_conference.tex
\documentclass{article} 
\usepackage{iclr2023_conference,times}
\iclrfinalcopy
\input{math_commands.tex}

\usepackage{hyperref}
\usepackage{url}

\title{Parameter-varying neural ordinary differential equations with partition-of-unity networks}

\author{Kookjin Lee  \\
School of Computing and Augmented Intelligence\\
Arizona State University\\
\texttt{kookjin.lee@asu.edu} \\
\And
Nathaniel Trask\\
Sandia National Laboratories\\
}

\usepackage{bm}
\usepackage{physics}
\usepackage{mathtools}
\mathtoolsset{showonlyrefs}
\usepackage{subfig}
\usepackage{amssymb,amsthm}
\usepackage{algorithmic}
\usepackage[linesnumbered,ruled,noend]{algorithm2e} 

\usepackage{multirow}
\usepackage{wrapfig}

\newtheorem{remark}{Remark}[section]

\newcommand{\Transpose}{^{\mathsf{T}}}

\newcommand{\npart}{n_{\text{part}}}
\newcommand{\npoly}{n_{\text{poly}}}

\begin{document}

\maketitle

\begin{abstract}
In this study, we propose parameter-varying neural ordinary differential equations (NODEs) where the evolution of model parameters is represented by partition-of-unity networks (POUNets), a mixture of experts architecture. The proposed variant of NODEs, synthesized with POUNets, learn a meshfree partition of space and represent the evolution of ODE parameters using sets of polynomials associated to each partition. We demonstrate the effectiveness of the proposed method for three important tasks: data-driven dynamics modeling of (1) hybrid systems, (2) switching linear dynamical systems, and (3) latent dynamics for dynamical systems with varying external forcing.
\end{abstract}

\section{Introduction}\label{sec:intro}

\subsection{Neural ordinary differential equations and their  variants}

Neural ordinary differential equations (NODEs) \citep{chen2018neural,weinan2017proposal,haber2017stable,lu2018beyond} are a class of continuous-depth neural network architectures that learn the dynamics of interest as a form of systems of ODEs: 
\begin{equation}\label{eq:nodes}
    \dv{\bm{h}(s)}{s} = \bm{f} (\bm{h}(s); \Theta),
\end{equation}
where $\bm{h}$ denotes a hidden state, $s$ represents a continuous depth, and the velocity function $\bm{f}$ is parameterized by a feed-forward neural network with learnable model parameters $\Theta$. 

As pointed out in \citep{massaroli2020dissecting}, the original NODE formulation \citep{chen2018neural}, is limited to incorporate the depth variable $s$ into dynamics as it is, e.g., by concatenating $s$ and $\bm{h}$, which are then fed to $\bm{f}(\bm{h}, s;\Theta)$, rather than constructing the map $s \mapsto \Theta(s)$. Recent studies investigate strategies to extend NODEs to be depth-variant. ANODEV2 \citep{zhang2019anodev2} proposes a hypernetwork-type approach which builds a coupled system of NODEs, where one NODE defines an evolution of state variables, while another NODE defines an evolution of model parameters. In \citep{massaroli2020dissecting}, stacked NODEs and Galerkin NODEs (GalNODEs) have been proposed where the evolution of model parameters are modeled as piecewise constants and a set of orthogonal basis, respectively. The idea of spectrally modeling model parameters has been further extended to enable basis transformation leading to stateful layers and compressible model parameters \citep{queiruga2021stateful}.

\begin{figure}[!h]
    \centering
    \subfloat[][Stacked NODE]{\includegraphics[width=.33\textwidth]{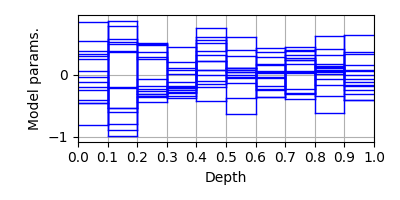} \label{fig:stackednode}}
    \subfloat[][Galerkin NODE]{\includegraphics[width=.33\textwidth]{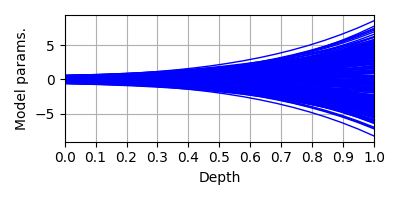} \label{fig:galnode}} 
    \subfloat[][POUNODE]{\includegraphics[width=.33\textwidth]{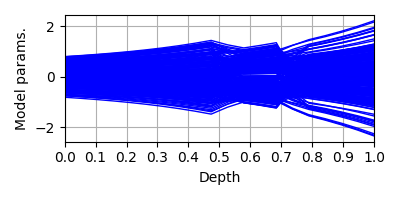} \label{fig:pounode}} 
    \caption{An illustrative example depicting evolution of model parameters for Stacked NODE, Galerkin NODE, and the proposed POUNODE. Models are trained to perform the binary classification task of two concentric circles.}
    \label{fig:toy_example}
\end{figure}

In this work, following the work by \citep{massaroli2020dissecting}  which has proposed two depth-variant NODEs: stacked NODEs (i.e., a piecewise constant representation of model parameters, e.g., Figure~\ref{fig:stackednode}) and Galerkin NODEs (i.e., spectral representation of model parameters, e.g., Figure~\ref{fig:galnode}). Inspired by these two variants, we propose an a combination of stacked and Galerkin NODEs leading to spectral-element-like \citep{patera1984spectral} or $hp$-finite-element-like \citep{solin2003higher} methods, which we denote by Partition-of-Unity NODEs (POUNODEs, e.g., Figure~\ref{fig:pounode}). We decompose the domain of model parameters (e.g., depth) into disjoint learnable partitions, with model parameters approximated on each as polynomials.

Our main contributions include 1) development of an $hp$-element-like method for representing the evolution of model parameters of NODEs and 2) to showcase the effectiveness of POUNODEs with three different important applications: learning hybrid systems, switching linear dynamics, and latent-dynamics modeling with varying external factor.   

\section{POUNets into NODEs}
We begin by introducing partition-of-unity networks (POUNets) \citep{lee2021partition}, which a particular type of deep neural network developed for approximating functions with exponential convergence. POUNets automatically learn partitions of the domain and simultaneously compute the coefficients of polynomials associated in each partition. Then we introduce a method to use POUNets for representing the evolution of model parameters for NODEs. 

\subsection{Partition-of-unity networks} 
Several recent works \citep{he2018relu,yarotsky2017error,yarotsky2018optimal,opschoor2019deep,daubechies2019nonlinear} on approximation theory of deep neural networks (DNNs) investigate the role of width and depth to the performance of DNNs and have theoretically proved the existence of model parameters of DNNs that emulate algebraic operations, a partition of unity (POU), and polynomials to exponential accuracy in the depth of the network. That is, in theory, with a sufficiently deep architecture, DNNs should be able to learn a spectrally convergent $hp$-element space by constructing a POU to localize polynomial approximation without a hand-tailored mesh. As has seen in \citep{fokina2019growing,adcock2020gap,lee2021partition}, however, such convergent behaviours in practice are not realized due to many reasons (e.g., gradient-descent-based training). In \citep{lee2021partition}, a novel neural network architecture, POUNets, has been proposed, which explicitly incorporates a POU and polynomial elements into a neural network architecture, leading to exponentially-convergent DNNs.

\begin{wrapfigure}{r}{0.4\textwidth}
  \vspace{-5mm}
  \begin{center}
  \subfloat[][Partitions]{\includegraphics[scale=.35]{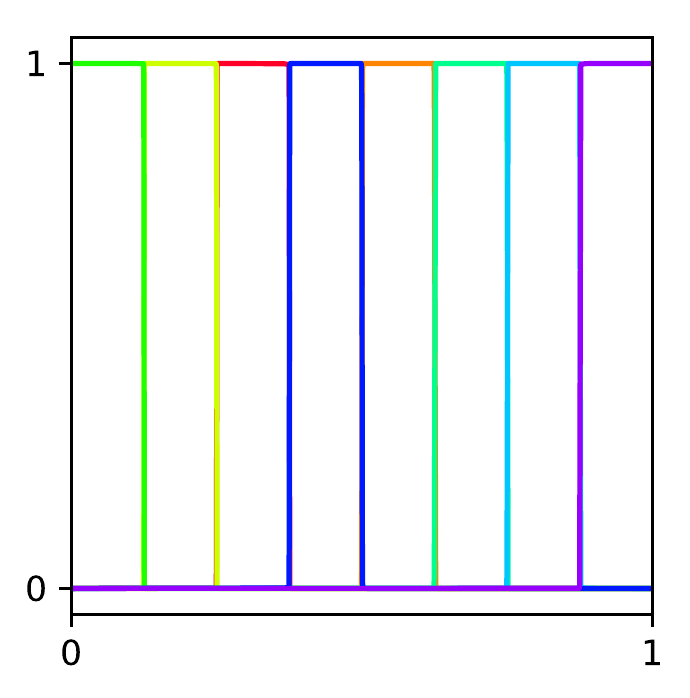} \label{fig:partitions_quad}}
  \subfloat[][Quadratic wave]{\includegraphics[scale=.35]{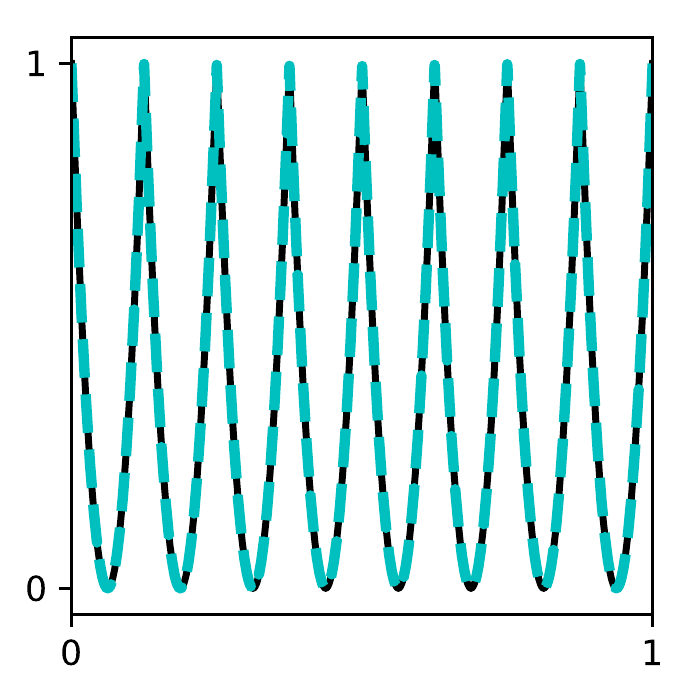} \label{fig:pred_quad}}    
  \end{center}
  \caption{Learned partitions (left) and predictions (cian dashed) depicted with the ground truth target function (black solid). 
  }\label{fig:pou_example}
  \vspace{-5mm}
\end{wrapfigure}

Mathematically, a POU can be defined as $\Phi(x) = \{\phi_i(x)\}_{i=1}^{\npart}$ satisfying $\sum_i \phi_i(x) = 1$ and $\phi_i \leq 0$ for all $x$. Then POUNets can be represented as
\begin{equation}
    y_{\text{POU}} (x) = \sum_{i=1}^{\npart} \phi_i(x;\pi) \sum_{j=1}^{\text{dim}(V)} \alpha_{i,j} \phi_j(x), 
\end{equation}
where $V = \text{span}(\{\psi_j\})$, typically taken as  the space of polynomials of order $m$, and $\Phi(x;\pi) = [\phi_1(x;\pi),\ldots, \phi_{\npart}(x;\pi)]$ is parameterized by a neural network with the model parameters $\pi$. To ensure the properties of the partition-of-unity, the output layer of the neural network $\Phi$ is designed to produce positive and normalized output (i.e., $\phi_i(x;\pi)\geq 0$ and $\sum_i \phi_i(x;\pi) = 1$). Figure~\ref{fig:pou_example} depicts an example of regressing a quadratic wave with a POUNet, where standard MLPs exhibit poor performance: the left panel shows the learned partitions and the right panel shows the ground truth target function (solid black) and the prediction (dashed cian). In each partition, a set of monomials with the maximal degree 2 is fitted optimally by solving local linear least-squares problems.

\subsection{Partition-of-unity-based neural ordinary differential equations}
Now, we introduce the proposed partition-of-unity-based neural ordinary differential equations, where the model parameters are represented as a POUNet:
$\Theta(\bm{s}) \in \mathbb{R}^{n_{\Theta}}$: 

\begin{equation}\label{eq:model_param}
    \Theta(\bm{s};\alpha,\pi) = \sum_{i=1}^{\npart} \phi_i(\bm{s};\pi) p_i(\bm{s}) =  \sum_{i=1}^{\npart} \phi_i(\bm{s};\pi) \sum_{j=1}^{\npoly}  \alpha_{i,j}  \psi_j(\bm{s}),
\end{equation}
where $\bm{s}$ denotes a set of variables whose domains are expected to have a set of partitions (e.g., $\bm{s}$ can be a depth variable in depth-continuous neural network architectures),   $\phi_i(\bm{s};\pi) \in \mathbb R$ denotes a partition of unity network, parameterized by $\pi$, $\psi_j(\bm{s}) \in \mathbb{R}$ denotes a polynomial basis, and $\alpha_{\cdot, j} \in \mathbb R^{n_{\Theta}}$ denote the polynomial coefficients. Thus, collectively,  there is a set of parameters $\alpha = (\alpha_1,\ldots,\alpha_{\npart})$ with $\alpha_i = [\alpha_{i,1} \cdots, \alpha_{i,\npoly}] \in \mathbb{R}^{ n_\theta \times  \npoly}$. In the following, we present a couple example cases of the types of the variables $\bm{s}$.

\paragraph{Temporally varying dynamics / depth variance} As in the typical settings of NODEs, when an MLP is considered to parameterize the velocity function, $\bm{f}(\cdot; \Theta)$, the model parameters can be represented as a set of constant-valued variables, $\Theta = \{(W_\ell, \bm{b}_\ell)\}_{\ell=1}^{L}$, where $W_\ell$ and $\bm{b}_\ell$ denote weights and biases of the $\ell$-th layer. As opposed to the depth-invariant NODE parameters $\Theta$, POUNODEs represent depth-variant NODEs (or non-autonomous dynamical systems) by setting the model parameters as 
\begin{equation}
    \Theta(t) = \{(W_\ell(t), \bm{b}_\ell(t))\}_{\ell=1}^{L},
\end{equation}
where $t$ denotes the time variable or the depth of the neural network and represent, and by representing $\Theta(t)$ as a POUNet as in Eq.~\eqref{eq:model_param} with $\bm{s}=t$. 

\paragraph{Spatially varying dynamics} Another example dynamical systems that can be represented by POUNODEs is a class of dynamical systems whose dynamics modes are defined differently on different spatial regions. In this case, the model parameters can be set as spatially-varying ones:
\begin{equation}
    \Theta(\bm{x}) = \{(W_\ell(\bm{x}), \bm{b}_\ell(\bm{x}))\}_{\ell=1}^{L}.
\end{equation}
and can be represented as a POUNet as in Eq.~\eqref{eq:model_param} with $\bm{s} = \bm{x}$.

\begin{remark}
Although not numerically tested in this study, the idea of representing the evolution of model parameters via POUNets can be applied to different neural network architectures, e.g., POU-Recurrent Neural Networks (POU-RNNs).
\end{remark}

\section{Use cases} 
This section exhibits example use cases where the benefits of using POUNODE can be pronounced. All implementations are based on \textsc{PyTorch} \citep{paszke2019pytorch} and the \textsc{TorchDiffeq} library \citep{chen2018neural} for the NODEs capability.

For all following experiments, we consider a POUNet, $\Phi = \{\phi_i \}_{i=1}^{\npart}$, based on a radial basis function (RBF) network \citep{broomhead1988radial,billings1995radial}; for each partition, there is an associated RBF layer, defined by its center and  shape parameter, and then the output of the RBF layers is normalized to satisfy the partition-of-unity property (refer to Appendix for more details).

\subsection{System identification of a hybrid system}\label{sec:hybrid}
As a first set of use cases, we apply POUNODEs for data-driven dynamics modeling. In particular, we aim to learn a dynamics model for a hybrid system, where the different dynamics models are mixed in a single system: a system consisting of multiple smooth dynamical flows (SDFs), each of which is interrupted by sudden changes (e.g., jump discontinuities or distributional shifts) \citep{van2000introduction}. 

Following \citep{shi2021segmenting},  we are interested in modeling a hybrid system, where external factors exist and results in sudden changes in the dynamics modes, which makes the applications of traditional dynamics modeling approach challenging. 

\begin{figure}[!ht]
    \centering
     \includegraphics[scale=.55]{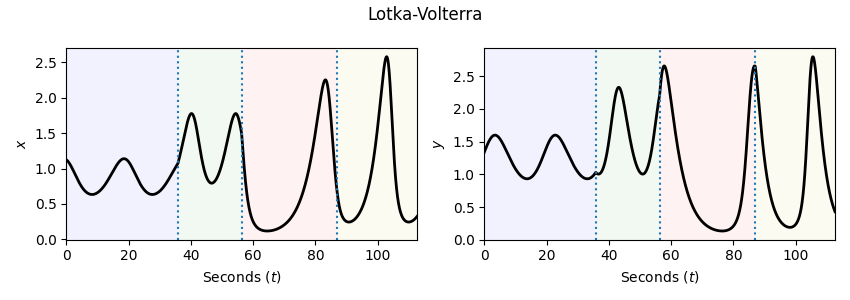} 
    \caption{A trajectory of four different dynamics generated from solving the LV equation.}
    \label{fig:lv_traj}
\end{figure}

Again, similar to \citep{shi2021segmenting}, as a benchmark, we consider the Lotka--Volterra (LV) equation:
\begin{align}
    \dot {x} &= a(t) x - b(t) xy, \\
    \dot {y} &= d(t) x y - c(t) y,
\end{align}
where $(a(t),b(t),c(t),d(t))$ are time-varying ODE parameters that define the dynamics. As a system identification benchmark problem, we generate a trajectory consisting of four different dynamics (SDFs) as depicted in Figure~\ref{fig:lv_traj}. The ODE parameters are chosen to be piecewise constant and the values of the parameters are listed in Table~\ref{tab:coeff_hybrid}. There are three change points at 35.85, 57.34, and 88.07 seconds, which are non-uniformly distributed over time. 

Now, we identify the system of the given trajectory using the proposed POUNODE. For the parameterization of the velocity function, we consider a dictionary-based approach: 
\begin{equation}\label{eq:dict}
    f_{\Theta(t)}(x) = \left(\Phi(x)\Transpose \Xi(t)\right)\Transpose,
\end{equation}
where $\Phi(x) \in \mathbb{R}^{p\times 1}$ denotes a vector of dictionaries and $\Xi(t) \in \mathbb{R}^{p\times n}$ denotes a trainable time-dependent coefficients (i.e., $\Theta(t) = \Xi(t)$). For the following experiments, we choose a set of polynomials as our dictionaries, i.e., $\Phi(x) =  [1, x, x^2, xy, y, y^2]\Transpose$ and $p=6$. The coefficients $\Theta(t)$ are modeled as a set of piecewise constant model parameters using POUNets such that
\begin{equation}
    \Theta(t;\alpha,\pi)  =  \sum_{i=1}^{\npart} \phi_i(t;\pi)   \left( \alpha_{i,1}  \psi_1(t)\right) = \sum_{i=1}^{\npart} \alpha_{i} \phi_i(t;\pi).
\end{equation}
That is, there is a set of constant coefficients associated with each partition, $\alpha_i \in \mathbb{R}^{p \times n}$, $i=1,\ldots,\npart$. Note that the above equation is a special case of the expression in Eq.~\eqref{eq:model_param} with $\npoly=1$.
\begin{table*}[h]
\centering
\resizebox{1\columnwidth}{!}{
\begin{tabular}{c|c|c|c|c}
    \hline
    $t$ (seconds) & [0, 35.85] & [35.86, 57.34] & [57.35,88.07] & [88.08,113.68]\\
    \hline
    \multirow{2}{*}{Ground truth} &  $\dot{x} = 0.3543 x - 0.2867 xy$ & $\dot{x} = 0.4301x - 0.2731 xy$ & $\dot{x} = 0.2500 xy - 0.2966 y$ & $\dot{x} = 0.3256 x - 0.3364 xy$\\
    & $\dot{y} = 0.3492xy - 0.3011y$ & $\dot{y} = 0.3847 xy -0.4695y$ & $\dot{y} = 0.3548 xy - 0.2568y$ & $\dot{y} = 0.4213xy - 0.4176y$\\
    \hline
    POUNODE & $\dot{x} = 0.3604x - 0.2895xy$ & $\dot{x} = 0.4334x - 0.2754xy$ & $\dot{x} = 0.2500x - 0.2950xy$ & $\dot{x} = 0.3285x-0.3384xy$\\
    ($\npart=4$) & $\dot{y} = 0.3447xy -0.2950 y$ & $\dot{y} = 0.3822xy - 0.4612y$ & $\dot{y} = 0.3532 xy - 0.2565y$ & $\dot{y} = 0.4134xy - 0.4110y$\\
    \hline
    POUNODE & $\dot{x} = 0.3530x - 0.2856 xy$ & $\dot{x} = 0.4353x - 0.2712xy $
    & $\dot{x} = 0.2508x - 0.2958xy$  & $\dot{x} = 0.3225x - 0.3342xy$\\
    ($\npart=8$) & $\dot{y} = 0.3512xy - 0.3007y$ & $\dot{y} = 0.3829xy - 0.4676y$ & $\dot{y} = 0.3561xy-0.2576y$ & $\dot{y} = 0.4213xy-0.4154y$\\
    \hline
\end{tabular}}
\caption{A hybrid Lotka--Volterra system consisting of four different dynamics. The coefficients for each dynamics of the ground truth system, and learned systems are listed.}
\label{tab:coeff_hybrid}
\end{table*}

For training, we use the training algorithm proposed in the work of sparse nonlinear dynamics identification method \citep{lee2021structure}. The essence is that a sparsity promoting L1 penalty (or L1 weight decay (LASSO) \citep{tibshirani1996regression}), is applied to the weight $\alpha = [\alpha_1, \ldots, \alpha_{\npart}]$ and an element of the weight whose magnitude is smaller than a certain threshold is pruned over the course of gradient-based training. We leave the details in Appendix. As we use the zero initialization (i.e., all elements of $\alpha_i$ are set to zero), we do not repeat the same experiments.  

\begin{figure}[!ht]
    \centering
     \includegraphics[scale=.45]{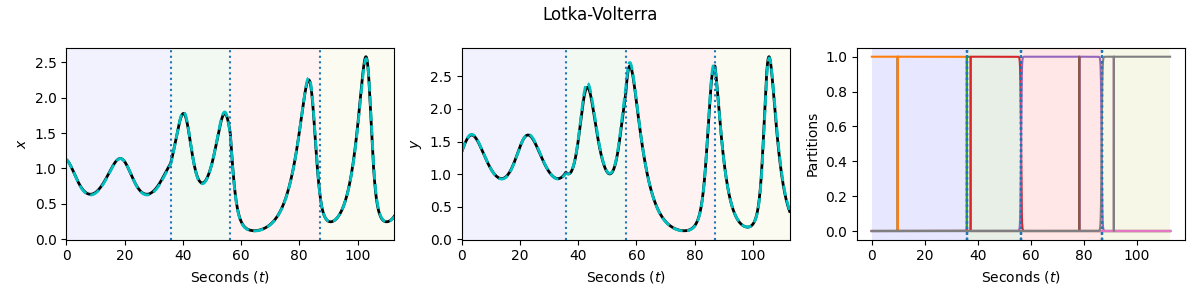} 
    \caption{A trajectory (dashed cyan color on two left panels) generated by solving learned dynamics model with eight beginning partitions ($\npart=8$) and the learned partitions (right).}
    \label{fig:lv_traj_pred}
\end{figure}

Table~\ref{tab:coeff_hybrid} reports the coefficients for the ground-truth systems (the second row) and the coefficients identified by using the proposed methods:  POUNODE ($\npart=4$), and POUNODE ($\npart=8$). POUNODE ($\npart=N)$ indicates that the model starts with $N$ partitions; some of them are expected to vanish as training proceeds, e.g.,  the right panel in Figure~\ref{fig:lv_traj_pred}. 
Figure~\ref{fig:lv_traj_pred} also depicts the trajectory of the learned dynamics (dashed cyan color on two left panels) that is almost overlapped with the ground-truth trajectory and the learned partitions (on the right panel).

\begin{wrapfigure}{r}{0.4\textwidth}
  \vspace{-6mm}
  \begin{center}
    \includegraphics[width=0.38\textwidth]{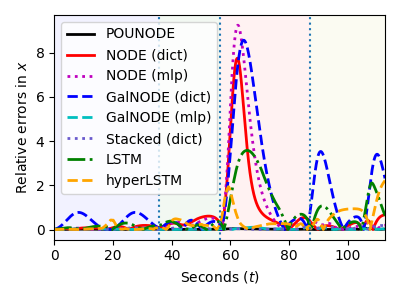}
  \end{center}
  \caption{Relative errors in predictions}\label{fig:lv_errors}
  \vspace{-3mm}
\end{wrapfigure}
Although the main objective of the system identification is to discover an interpretable model (via imposing strong inductive biases such as the choice of dictionaries), we can also see how well the model fits to the data. As a baseline for comparison, we test RNN-LSTM, RNN-GRU,  hyperLSTM \citep{ha2017hyper}, NODE ($\npart=1$, $\npoly=1$), StackedNODE,  GalNODE ($\npart=1$, $\npoly=3$), and ANODEv2\citep{gholami2019anode}\footnote{The results of RNN-GRU and ANODEv2 were not reported as the both models did not seem to be trained well under the experimental configuration used for training the proposed method and other baselines.}. Here, we employ the same dictionary-based parameterization (Eq.~\eqref{eq:dict}) for NODE, StackedNODE and GalNODE  as well as  the ``black-box'' MLP parameterization for NODE and GalNODE. For MLP, we consider 4 layers with 25 neurons in each layer. Figure \ref{fig:lv_errors} shows the time-instantaneous relative error of the trajectory of $x(t)$, i.e., $e(t) = \frac{| x(t) - \tilde x(t) |}{|x(t)|}$, where $\tilde x(t)$ denotes the predictions and $|\cdot|$ denotes an absolute value. As Figure~\ref{fig:lv_errors} shows POUNODE outperforms other baseline approaches in terms of accuracy and have comparable accuracy with Stacked NODE (dictionary-based, 8 partitions) and GalNODE (mlp): the relative errors measured in L2-norm are 0.0160, 0.0615, and 0.0264 for POUNODE, StackedNODE  and GalNODE, respectively. StackedNODE, however, is not capable of pinpointing the change points, and GalNODE requires a much larger number of model parameters ($\times 50$ more parameters, compared to POUNODE). 

\subsection{Switching linear dynamical systems}\label{sec:slds}
\begin{figure}[!ht]
    \centering
    \subfloat[][Ground-truth]{\includegraphics[scale=.33]{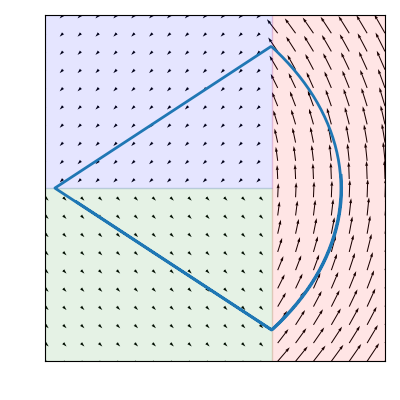} \label{fig:groundtruth}}
    \subfloat[][Initial partitions]{\includegraphics[scale=.33]{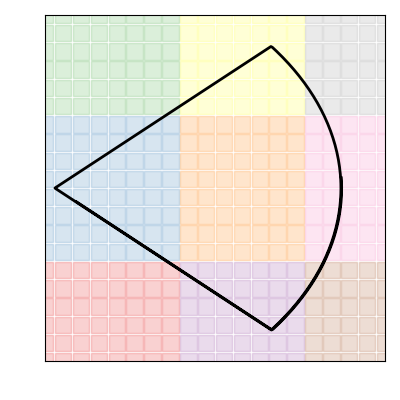} \label{fig:slds_initpart}} 
    \subfloat[][POUNODE]{\includegraphics[scale=.33]{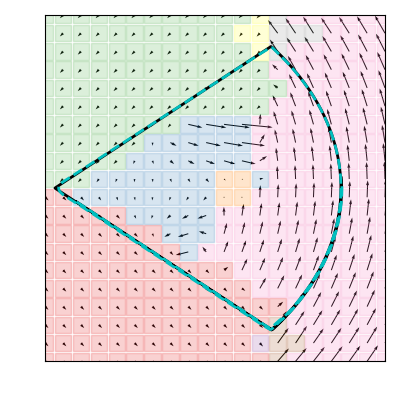} \label{fig:slds_learnedpart}} 
    \subfloat[][Trajectories over time]{\includegraphics[scale=.35]{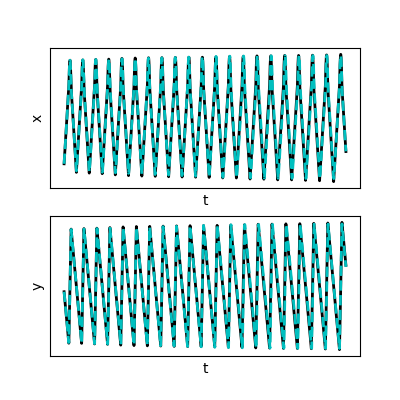} \label{fig:slds_trajs}}
    \caption{A ground-truth trajectory with the ground-truth vector field (left), a computed trajectory computed from the learned vector field (right). Different colors indicate the different partitions.}
    \label{fig:slds_traj}
\end{figure}

As a next set of use cases, we consider switching linear dynamical systems (SLDS) consisting of multiple sequences of simple dynamical modes that change the dynamics mode based on a discrete switch \citep{ackerson1970state, chang1978state,ghahramani1996switching,fox2008nonparametric,linderman2016recurrent}. We are interested in data-driven dynamics modeling of SLDS in the continuous-time setting as has considered in \citep{chen2020learning}, taking the ground-truth dynamics as described in Table~\ref{tab:coeff_slds}. At the boundary of each spatial subdomain (as depicted in Figure~\ref{fig:groundtruth}), the dynamics changes instantaneously and, thus, the resulting dynamics consists of sequences of different dynamics and can exhibit discontinuities at the moment of switching.   %

This benchmark problem considered is an SLDS example of a particle moving around a fan-shaped synthetic race track as in \citep{chen2020learning}, which has been originally adapted from \citep{linderman2016recurrent}. Figure~\ref{fig:slds_traj} (left) depicts an example of the ground-truth trajectory and the vector field and the analytical expression of the ODEs can be found in Table~\ref{tab:coeff_slds}. 

Our goal is to utilize POUNODE to model an SLDS by treating SLDS parameters as parameters that are dependent on the spatial coordinates $\bm{x}$. That is, we consider a time-continuous model 
\begin{equation}
    \dv{\bm{x}}{t} = \bm{f}_{\Theta(\bm{x})}(\bm{x}) = C(\bm{x}) \bm{x} + \bm{d}(\bm{x}),
\end{equation}
where $C(\bm{x})$ and $\bm{d}(\bm{x})$ are model parameters, $\Theta(\bm{x}) = [C(\bm{x}),  \bm{d}(\bm{x})] \in \mathbb{R}^{2\times 3}$, that are piecewise constant on each partition:
\begin{equation}
    \Theta(\bm{x};\alpha,\pi) = \sum_{i=1}^{\npart} \alpha_{i} \phi_i(\bm{x};\pi), 
\end{equation}
where $\alpha_i \in \mathbb{R}^{2\times 3}$ denotes the $i$-th coefficients defined on the $i$-th partition, $\phi_i$. Our intention is to learn the three disjoint spatial regions as disjoint partitions and the associated piece-wise constant coefficients to correctly identify the vector field. 

For training, we again use the same algorithm, proposed in \citep{lee2021structure}, which we summarize in Appendix. For the system identification task, we use a single trajectory to train the model.

\begin{table*}[t]
\centering
\resizebox{1\columnwidth}{!}{
\begin{tabular}{c|c|c|c}
    \hline
    Coordinates ($x,y$) & $x<2,y<0 $ & $x<2,y \geq 0$ & $x\geq 2$ \\
    \hline
    \multirow{2}{*}{Ground truth} &  $\dot{x} = 1$ & $\dot{x} = -1$ & $\dot{x} = -y$ \\
    & $\dot{y} = -1$ & $\dot{y} = -1$ & $\dot{y} = x + 2$ \\
    \hline
    \hline
     Partitions & Top-left (green) partition & Bottom-left (red) partition & Right (pink) partition \\
    \hline
    \multirow{2}{*}{POUNODE} & $\dot{x} = 0.9980$ & $\dot{x} = -1.0033$ & $\dot{x} = -1.0000 y$\\
     & $\dot{y} = -1.0000$ & $\dot{y} = -0.9965$ & $\dot{y} = 0.9977x + 1.9968 $ \\
    \hline
\end{tabular}}
\caption{A switching linear dynamical system:  The ODEs of each dynamics of the ground truth system, and learned systems are listed.}
\label{tab:coeff_slds}
\end{table*}

Figures~\ref{fig:slds_initpart}--\ref{fig:slds_trajs} show the initial $3\times 3$ partitions (Figure~\ref{fig:slds_initpart}), the learned partitions and the trajectory produced by solving the learned dynamics model (Figure~\ref{fig:slds_learnedpart}), and the trajectories of each state variable (Figure~\ref{fig:slds_trajs}). Table~\ref{tab:coeff_slds} reports the identified systems in each region. As reported in Figures~\ref{fig:slds_initpart}--\ref{fig:slds_trajs} and Table~\ref{tab:coeff_slds}, POUNODE successfully identify the benchmark SLDS with the errors in the third/fourth most significant digits.

\begin{figure}[!htb]
    \centering
    \subfloat[][Initial partitions]{\includegraphics[scale=.33]{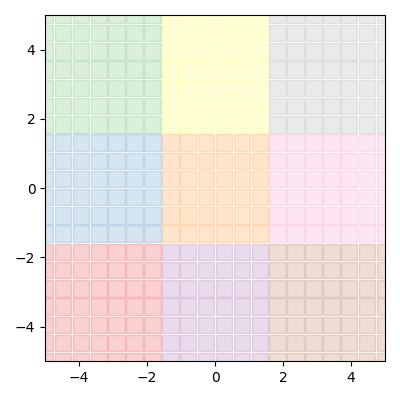} \label{fig:slds_initpart_multi}} 
    \subfloat[][POUNODE]{\includegraphics[scale=.33]{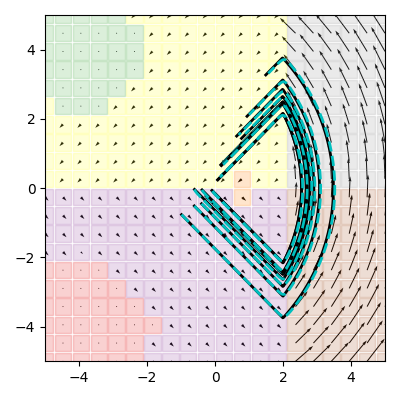} \label{fig:slds_learnedpart_multi}} 
    \subfloat[][Test trajectory \#1]{\includegraphics[scale=.36]{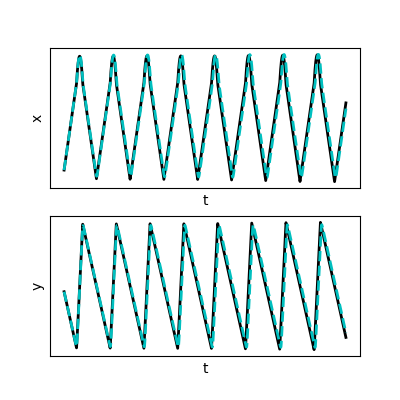} \label{fig:slds_trajs_multi1}}
    \subfloat[][Test trajectory \#2]{\includegraphics[scale=.36]{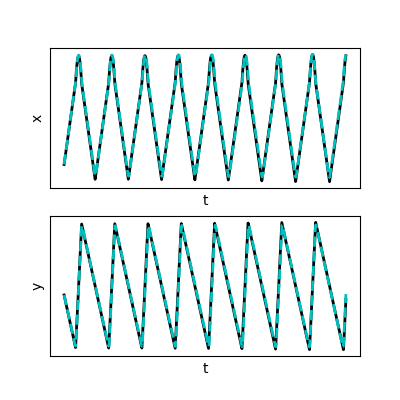} \label{fig:slds_trajs_multi2}}
    \caption{Ground-truth trajectories (solid black) and trajectories computed from the learned vector field (dashed cyan) are depicted. Different colors in background indicate the different partitions: initial partitions (Figure~\ref{fig:slds_initpart_multi}) and learned partitions (Figure~\ref{fig:slds_learnedpart_multi}).}
    \label{fig:slds_traj_multi}
\end{figure}

Figure~\ref{fig:slds_traj_multi} reports the results of learning dynamics using multiple trajectories and applying the learned dynamics in the predictive setting. For this experiments, we have generated 80 training, 10 validation, and 10 test trajectories with varying initial conditions. Figures~\ref{fig:slds_learnedpart_multi}--\ref{fig:slds_trajs_multi2} depict the ground-truth trajectories (solid black) and the trajectories computed from the learned dynamics model (dashed cyan). Figure~\ref{fig:slds_learnedpart_multi} shows that there are four remaining partitions, where the learned coefficients are as follows:
\begin{equation}
\begin{array}{cccc}
\text{(purple partition)}     &  \text{(yellow partition)} & \text{(gray partition)} & \text{(brown partition)}\\
\hline
\dot{x}= 0.9992,      & \dot{x}=-0.9997, & \dot{x} = -0.9995 y, & \dot{x} = -0.9985 y,\\
\dot{y}= -0.9982,     & \dot{y} = -1.0010,  & \dot{y} = 0.9966x + 2.0034,  & \dot{y} = 0.9962x + 2.0217.
\end{array}
\end{equation}

Compared to the approach where the method learns a differential event function \citep{chen2020learning}, the proposed approach directly learns the vector fields that are differently defined in each spatial domain and, thus, does not require to specify in advance how many events of switching dynamics will happen in the simulation run. 

\subsection{Latent dynamics modeling (reduced-order modeling)}\label{sec:rom}
The next use case is a latent-dynamics modeling in the context of reduced-order modeling (ROM), a computational framework that is widely investigated in the field of computational science and engineering \citep{fulton2019latent, lee2020model, lee2021deep}.  The main goal of developing ROMs is to provide a means to perform rapid simulations of complex physical phenomena (typically described in partial differential equations) to support time-critical applications such as control. 

As elaborated in \citep{lee2021deep}, a latent-dynamics modeling requires two main components: 1) an \textit{embedding}, i.e., a nonlinear mapping between  high-dimensional dynamical-system states and low-dimensional latent states, and (2) a dynamics model, i.e., the time evolution model of the latent states. For learning an embedding, nearly all traditional numerical methods seek a linear embedding, which is typically defined by principal component analysis, or ``proper orthogonal decomposition'' (POD) \citep{holmes2012turbulence}, performed on
measurements of the high-dimensional states. Recent approaches, on the other hand, explore the use of deep neural networks, (autoencoders \citep{hinton2006reducing}, in particular), to build a nonlinear embedding \citep{morton2018deep, wiewel2019latent, fulton2019latent, lee2020model, lee2021deep}. After learning the embedding, a (nonlinear) latent-dynamics model is constructed, representatively, via long short-term memory \citep{hochreiter1997long}, Koopman operators \citep{li2019learning,azencot2020forecasting}, and NODEs \citep{chen2018neural, lee2020parameterized}.

\begin{wrapfigure}{r}{0.4\textwidth}
  \vspace{-4mm}
  \begin{center}
    \includegraphics[width=0.38\textwidth]{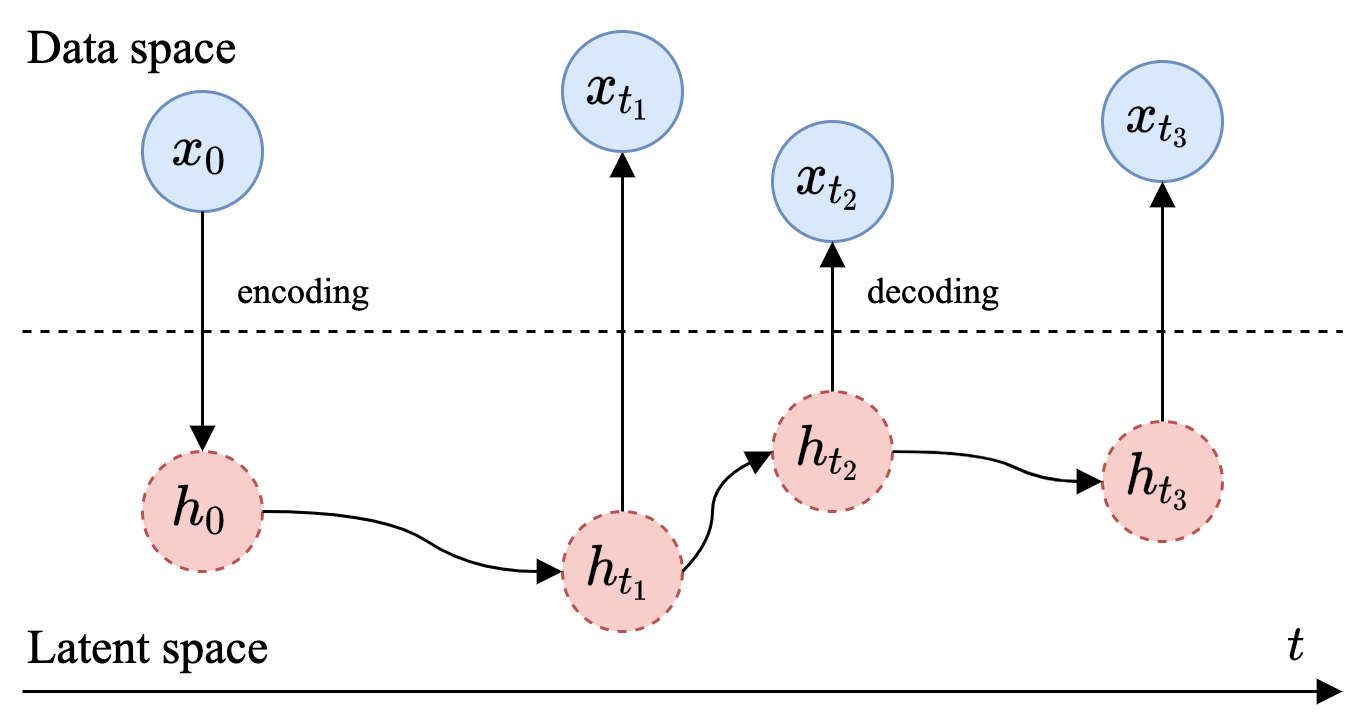}
  \end{center}
  \caption{A latent-dynamics model}\label{fig:latent_dyn}
  \vspace{-4mm}
\end{wrapfigure}
In the following experiment, we choose a linear embedding, defined by a POD basis matrix, $\varphi \in \mathbb{R}^{N \times p}$, where $N$ and $p$ denote the dimensions of the high dimensional space and the latent space. The encoding and the decoding are defined as $\bm h = \varphi \bm x$ and $\bm x = \varphi\Transpose \bm h$, where $\bm x\in \mathbb R^{N}$ and $\bm h \in \mathbb R^{p}$. Given the linear encoder and the decoder, we learn the latent dynamics with the proposed POUNODE. As Figure \ref{fig:latent_dyn} illustrates that an initial high-dimensional state is encoded into the latent initial state, future latent states are computed via the forward pass of POUNODE, and the high-dimensional approximate states are computed via the decoder.

As a benchmark problem, we consider 1-dimensional inviscid Burgers' equation with a parameterized forcing term; setting different values to the parameter change the dynamics. We generate a  35-seconds-long trajectory that has two change points at $t=[13.4 , 22.1]$ seconds, where the value of the forcing parameter changes. 
We test the latent-dynamics modeling in a reconstructive setting with a single trajectory and we set the original data dimension to be $N=256$ and the latent dimension to be $p=3$.

The velocity function is parameterized as an MLP with 2 hidden layers, 25 neurons in each layer, and the hyperbolic Tangent nonlinearity such that $\bm{f} = W^{(3)} \sigma(W^{(2)} \sigma(W^{(1)} \bm{h} + \bm b^{(1)}) + \bm b^{(2)})+\bm b^{(3)}$, where the weights and the biases are function of time  $\Theta(t) = \{(W^{(\ell)}(t), \bm{b}^{(\ell)}(t) \}_{\ell=1}^{3}$, which are modeled as a POUNet (Eq.~\eqref{eq:model_param}). 

\begin{wraptable}{r}{0.45\textwidth}
\caption{Performance}\label{tab:rom}
\resizebox{.45\textwidth}{!}{
\begin{tabular}{cc}\\
\hline
Models & Accuracy \\
\hline
RNN-LSTM & 0.4367 $\pm$ 0.08620\\ \hline
RNN-GRU &  0.3035 $\pm$ 0.06128\\ \hline
HyperLSTM \citep{ha2017hyper} & 0.2244 $\pm$ 0.06812\\ \hline
NODE ($\npart=1, \npoly=1$) & 0.2981 $\pm$ 0.00275\\   \hline
ANODEv2 \citep{gholami2019anode} & 0.3589 $\pm$ 0.08057\\   \hline
GalNODE ($\npart=1, \npoly=3$) & 0.4783 $\pm$ 0.15414 \\  \hline
StackedNODE ($\npart=6$, fixed) & 0.3659 $\pm$ 0.00472 \\  \hline
POUNODE ($\npart=3, \npoly=1$) & 0.1147 $\pm$ 0.00072 \\  \hline
POUNODE ($\npart=6, \npoly=1$) & 0.0731 $\pm$ 0.00286\\  \hline
POUNODE ($\npart=9, \npoly=1$) & 0.0730 $\pm$ 0.00057 \\  
\hline
\end{tabular}
}
\end{wraptable} 
Table~\ref{tab:rom} reports relative errors in the $L2$-norm, $\frac{\| X - \tilde X \|_{\mathsf F}}{\| X \|_{\mathsf F}}$, where $X, \tilde X \in \mathbb{R}^{N \times n_{\text{seq}}}$ denote the ground-truth solution measurements and the predicted solutions. POUNODE with $\npart=1$ and $\npoly$ is equivalent to NODE and POUNODE with $\npart=1$ and $\npoly=3$ is conceptually same as the GalNODE. We also observed that setting $\npart \geq 6$ does not improve the performance significantly. As baselines of comparisons, we assess the performance of RNN-LSTM, RNN-GRU, HyperLSTM \citep{ha2017hyper}, ANODEv2 \citep{gholami2019anode}, and Stacked NDOEs (with 6 fixed partitions), of which results are reported in Table \ref{tab:rom}. For each model, we repeat perform 5 runs of experiments with different random seeds. The details of the neural network architecture and hyperparameter choices are in Appendix.

\begin{figure}[h]
    \centering
    \subfloat[][$\npart=1$]{\includegraphics[scale=.3]{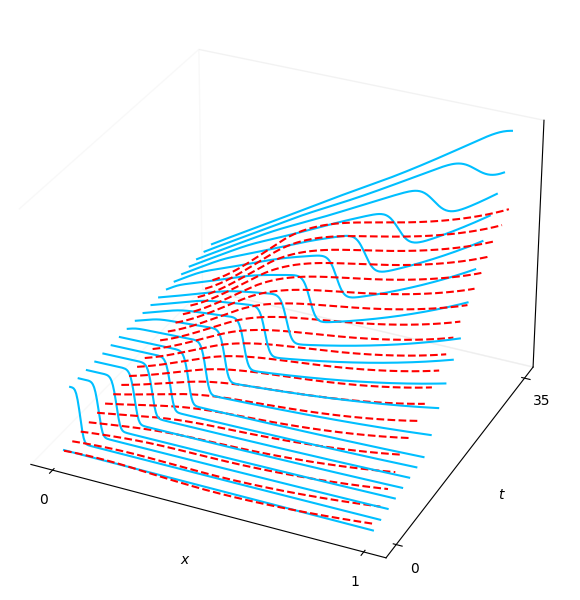} }
    \subfloat[][$\npart=3$]{\includegraphics[scale=.3]{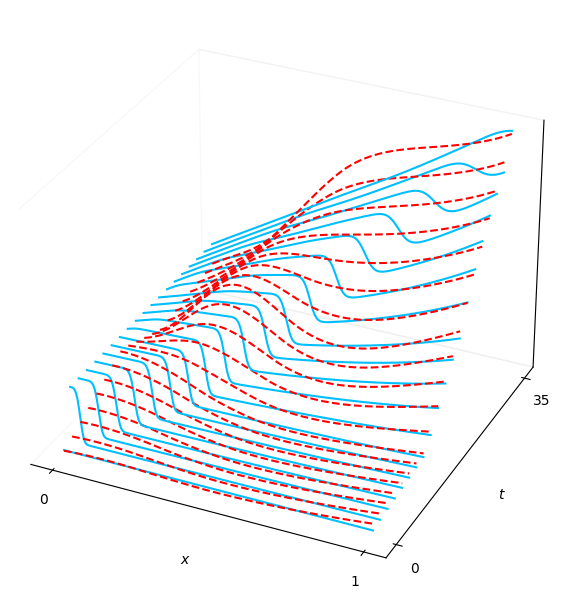} }
    \subfloat[][$\npart=9$]{\includegraphics[scale=.3]{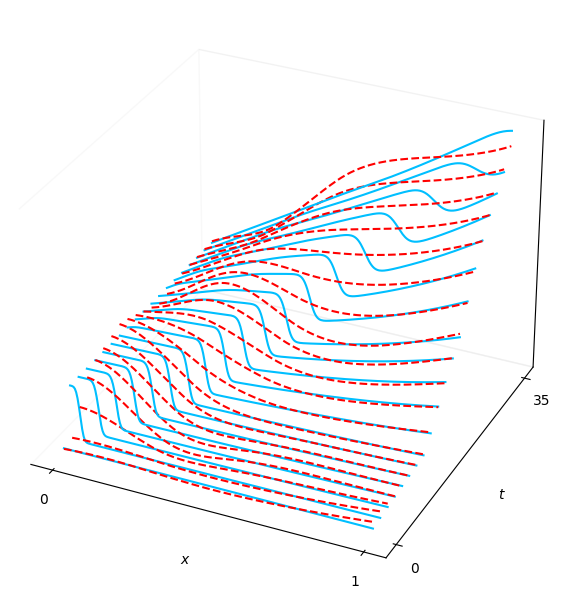} }
    \caption{The ground-truth solution snapshots (solid blue) and the approximated solution snapshots (dashed red) for varying latent dynamics models with $\npart = \{1,3,9\}$.}
    \label{fig:rom_pred}
\end{figure}

Figure~\ref{fig:rom_part} illustrates the ground truth change points (the black dashed vertical lines), where in between the forcing term remain the same (the regions highlighted with different colors), and the learned partitions. The partitions are learned to have disjoint sections that do not cross over the change points, and some of the unnecessary partitions are eliminated. Figure~\ref{fig:rom_pred} depicts the ground-truth solution snapshots (solid blue) and the approximated solution snapshots (dashed red) for varying latent dynamics models with $\npart = \{1,3,9\}$. The approximate solutions are smooth as they are represented as linear combinations of three principal basis ($\varphi \in \mathbb{R}^{N \times p}$ with $p=3$), however the approximate solutions that are generated with the latent dynamics models ($\npart\geq 3$) shows that they can match the shock locations (i.e., a place of the discontinuity in each snapshot). 

\begin{figure}[h]
    \centering
    \subfloat[][$\npart=3$]{\includegraphics[scale=.6]{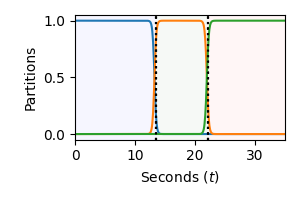} }
    \subfloat[][$\npart=6$]{\includegraphics[scale=.6]{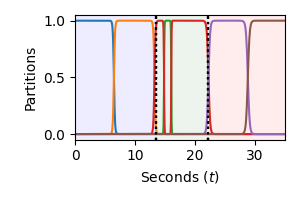} }
    \subfloat[][$\npart=9$]{\includegraphics[scale=.6]{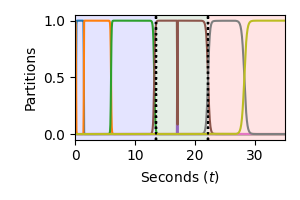} \label{fig:rom_part_c}}
    \caption{Learned partitions for the latent-dynamics models. The subcaption, $\npart=k$, indicates the number of the beginning partitions.}
    \label{fig:rom_part}
\end{figure}

\section{Discussion}
\subsection{Limitations and future directions} 

\paragraph{Minibatching} Minibatching trajectories with different change points requires multiple POUNets, where each POUNet needs to be a realization of an input-data dependent POUNet, i.e., $\Theta(\bm{s}, \bm{x}^{\text{(input)}} ;  \alpha,\pi)$, where $\bm{x}^{\text{(input)}}$ denotes the input data. An approach similar to data-controlled NODEs proposed in \citep{massaroli2020dissecting} can be extended to be equipped with POUNets such that 
$\dv{\bm{h}(t)}{t}~=~ \bm{f}(\bm{h}(t),\bm{x}^{\text{(input)}};\Theta(\bm{s}, \bm{x}^{\text{(input)}}))$.

\paragraph{Predictive tasks} As shown in Sections~\ref{sec:hybrid} and \ref{sec:rom}, the proposed POUNODEs has demonstrated their effectiveness for identifying or building a surrogate model for a hybrid system. As demonstrated in \citep{shi2021segmenting}, the proposed models can be used for extrapolation in a somewhat limited scenario, where the dynamical mode remain unchanged. To be useful in the extrapolation settings, where the future dynamics is under temporal drift, alternative approaches that learn to produce future model parameters based on past sets of model parameters (e.g., hypernetwork-based approaches \citep{ha2017hyper}) or that exploit hierarchical structures of time-series for forecasting (e.g., N-BEATS \citep{oreshkin2019n}).

\paragraph{Fast optimizer} In the original work of POUNets  \citep{lee2021partition}, a fast optimizer, which alternates between gradient descent updates for updating partition parameters and least-squares solves for computing optimal polynomial coefficients, has been proposed for solving polynomial regression problems and has demonstrated the faster convergence. Thus, as opposed to the gradient-descent-based optimizer used in this work, which updates all model parameters simultaneously, developing an optimizer tailored to POUNODE would allow faster and more accurate training. 

\paragraph{POUNODEs as general NN architectures} As in previous work \citep{zhang2019anodev2,massaroli2020dissecting}, the depth-variant neural ODEs have demonstrated increased performance in other downstream tasks, e.g., image classification. We have tested POUNODEs, where convolutional kernels are spectrally represented, for image classification with CIFAR-10 by using the same setting considered in \citep{dupont2019anode, massaroli2020dissecting} and observe only marginal improvements (1$\sim$2\% increase in the test accuracy, but with the increase in the number of function evaluations). We expect that the benefits of using POUNODEs can be more pronounced in more complex settings, e.g., replacing multiple ResBlocks in ResNet-151 \citep{he2016deep} with a small number of POUNODE-Blocks, and plan to further investigate the performance of POUNODEs in those settings. 

\section{Conclusion}
In this study, we have introduced a new variant of NODEs (POUNODEs) with evolving model parameters, where the evolution is modeled by using partition-of-unity networks. We have demonstrated the effective of the proposed POUNODEs with three important case studies: learning hybrid dynamical systems, switching linear dynamics, and latent dynamics modeling with varying external factors. In those use-cases, we have demonstrated that the POUNODEs are very effective and outperform the baselines including the previous depth-variant NODEs and hypernetwork-based LSTMs. 


\section{Acknowledgements}
This paper describes objective technical results and analysis. Any subjective views or opinions that might be expressed in the paper do not necessarily represent the views of the U.S. Department of Energy or the United States Government. Sandia National Laboratories is a multimission laboratory managed and operated by National Technology \& Engineering Solutions of Sandia, LLC, a wholly owned subsidiary of Honeywell International Inc., for the U.S. Department of Energy’s National Nuclear Security Administration under contract DE-NA0003525.

\bibliography{iclr2023_conference}
\bibliographystyle{iclr2023_conference}

\appendix
\section{Training algorithms}
For the system identification, where the dictionary-based parameterization of the right-hand side of ODEs is employed (in Sections~\ref{sec:hybrid}--\ref{sec:slds}), we use the neural ODE-based sparse nonlinear dynamics identification method (i.e., nerual SINDy)  developed in \citep{lee2021structure}.

During the training, the model takes the forward pass by solving initial value problems as in neural ODEs. Then, as a training objective, the $L$1-distance between the data and the prediction is minimized. In addition, to promote the sparsity of the coefficients matrix $\Xi$, the elements of $\Xi$ is penalized with the $L$1-penalty:
\begin{equation}
    L = \frac{1}{n_b  n_\text{seq}} \sum_{j=1}^{n_b}\sum_{i=1}^{n_{\text{seq}}} \left \vert \bm{x}_i^{(j)} - \tilde{\bm{x}}_i^{(j)} \right \vert + \lambda \| \Xi \|_1,
\end{equation}
where $n_b$ and $n_{\text{seq}}$ denote the size of a minibatch and the length of sequences in the minibatch, $\bm{x}$ and $\tilde{\bm{x}}$ denote the data and the prediction, and $\lambda$ is the penalty weight, which is set as $10^{-4}$. 

In addition, over the course of training, neural SINDy prunes the coefficients based on their absolute magnitude with a certain threshold, $\tau$:
\begin{equation}
    [\Xi]_{kl} = 0 \quad \text{if} \quad \vert [\Xi]_{kl} \vert < \tau.
\end{equation}
We set $\tau = 10^{-6}$ for all experiments. Pruning is applied to learned $\Xi$ every feeding 100 minibatches. 

\section{Radial-basis-function-based POUNets}
For all experiments, we use the simple RBF-based POUNets as follows (in a one-dimension case):
\begin{equation}
    \phi_i(x) = \frac{\exp\left( - \frac{| x - c^{(i)} |}{b^{(i)}} \right)}{ \sum_k \exp\left( - \frac{| x - c^{(k)} |}{b^{(k)}} \right)},
\end{equation}
where $\{(c^{(i)}, b^{(i)} \}_{i=1}^{\npart}$ is a set of learnable parameters. Here,  $c^{(i)}$ and $b^{(i)}$ denote the center and the bandwidth of the RBF, respectively.

The centers are initialized to be on a uniform grid of the spatial domain.

\section{Model architectures and Hyperparameters}

\paragraph{System identification of a hybrid system experiments}
\begin{itemize}
    \item Model architectures
    \begin{itemize}
        \item (Dictionary-based) NODE, StackedNODE, GalNODE, POUNODE: a single layer that linearly combines the output of the dictionaries $\Phi(x) =  [1, x, x^2, xy, y, y^2]$
        \item (MLP-based) NODE, GalNODE : 4 layers with 25 neurons and Tanh activation
        \item LSTM: 4 stacked LSTM cells with 25 neurons for hidden and cell states
        \item hyperLSTM: 4 stacked LSTM and hyperLSTM cells with 25 neurons for hidden, cell states and 25 neurons for hyper and embedding units
    \end{itemize}
    \item Hyperparameters
    \begin{itemize}
        \item Learning rate: 0.01
        \item Max epoch: 3000
        \item Batch size: 1 (50 for hyperLSTM)
        \item Batched subsequence length: 100
        \item ODE integrator: Dormand--Prince (dopri5) \citep{dormand1980family} with the relative tolerance $10^{-7}$ and the absolute tolerance $10^{-9}$
    \end{itemize}
\end{itemize}
\paragraph{Learning switching linear dynamical systems}
\begin{itemize}
    \item Model architectures
    \begin{itemize}
        \item POUNODE: a single layer that linearly combines the output of dictionaries,  $\Phi(x) =  [1, x, y]$
    \end{itemize}
    \item Hyperparameters
    \begin{itemize}
        \item Learning rate: 0.01
        \item Max epoch: 3000
        \item Batch size: 1 
        \item Batched subsequence length: 100
        \item ODE integrator: Runge--Kutta of order 4
    \end{itemize}
\end{itemize}
\paragraph{Latent dynamics modeling}
\begin{itemize}
    \item Model architectures
    \begin{itemize}
        \item (MLP-based) NODE, StackedNODE, GalNODE, POUNODE : 2 layers with 25 neurons and Tanh activation
        \item ANODEv2: 2 layers with 24 neurons and Tanh for the main NODEs, 2 layers with 50 neurons and Tanh for the weight NODEs  
        \item LSTM: 2 stacked LSTM cells with 25 neurons for hidden and cell states
        \item hyperLSTM: 2 stacked LSTM and hyperLSTM cells with 25 neurons for hidden, cell states and 50 neurons for hyper and embedding units
    \end{itemize}
    \item Hyperparameters
    \begin{itemize}
        \item Learning rate: 0.01
        \item Max epoch: 500
        \item Batch size: 1 
        \item Batched subsequence length: 50
        \item ODE integrator: Dormand--Prince (dopri5) \citep{dormand1980family} with the relative tolerance $10^{-7}$ and the absolute tolerance $10^{-9}$
    \end{itemize}
\end{itemize}

\section{1D inviscid Burgers' equation}
As a benchmark problem for reduced-order modeling (shown in Section~\ref{sec:rom} latent-dynamics modeling), we consider 1-dimensional inviscid Burgers' equation, which is defined as, 
\begin{equation}
    \begin{split}
    \pdv{w(x,t;\mu)}{t} + \pdv{f(w(x,t;\mu))}{x} &= 0.02 e ^{\mu x}, \qquad \forall x \in [0, 100], \: \forall t \in [0, T]\\
    w(0,t;\mu) &= 4.5, \qquad \forall t \in [0,T]\\
    w(x,0) &= 1, \qquad \forall x \in [0, 100],
    \end{split}
\end{equation}
where $\mu$ defines the the forcing term, and we set $\mu = \mu(t)$ to be a time dependent function. In the high-fidelity simulation, we set $\mu(t)$ to be a piecewise-constant function over time such that 
\begin{equation}
    \mu(t) = \left\{ \begin{array}{l l}
    0.005 & \text{if } t \leq 13.4\\
    0.015 & \text{if } 13.4 < t \leq 22.2\\
    0.025 & \text{if } t > 22.2
    \end{array}
    \right.
\end{equation}
For the discretization, we apply Godunov's scheme with 256 control volumes (i.e., $N=256$ degrees of freedom) and the backward-Euler scheme with 600 uniform time steps.  

\end{document}

%% file: math_commands.tex
\usepackage{amsmath,amsfonts,bm}

\def\eqref#1{equation~\ref{#1}}

\def\1{\bm{1}}

\DeclareMathAlphabet{\mathsfit}{\encodingdefault}{\sfdefault}{m}{sl}
\SetMathAlphabet{\mathsfit}{bold}{\encodingdefault}{\sfdefault}{bx}{n}

